\pdfoutput=1

\documentclass[11pt]{article}

\usepackage{naacl2021}

\usepackage{graphicx}

\usepackage{balance}

\usepackage{times}
\usepackage{latexsym}

\usepackage[T1]{fontenc}

\usepackage[utf8]{inputenc}

\usepackage{amsmath}
\DeclareMathOperator*{\argmax}{arg\,max}

\usepackage{microtype}

\title{When and Why does a Model Fail? A Human-in-the-loop Error Detection Framework for Sentiment Analysis}

\author{Zhe Liu \\
  IBM Research - Almaden \\
  San Jose, CA, USA  \\
  \texttt{liuzh@us.ibm.com} \\\And
  Yufan Guo \thanks{YG was affiliated with IBM Research - Almaden at the time of the work reported in this paper.} \\
  Amazon Alexa AI \\
  Seattle, WA, USA \\
  \texttt{gyufan@amazon.com} \\\And
  Jalal Mahmud \\
  IBM Research - Almaden \\
  San Jose, CA, USA  \\
  \texttt{jumahmud@us.ibm.com} \\}

\begin{document}
\maketitle
\begin{abstract}
Although deep neural networks have been widely employed and proven effective in sentiment analysis tasks, it remains challenging for model developers to assess their models for erroneous predictions that might exist prior to deployment. Once deployed, emergent errors can be hard to identify in prediction run-time and impossible to trace back to their sources. To address such gaps, in this paper we propose an error detection framework for sentiment analysis based on explainable features. We perform global-level feature validation with human-in-the-loop assessment, followed by an integration of global and local-level feature contribution analysis. Experimental results show that, given limited human-in-the-loop intervention, our method is able to identify erroneous model predictions on unseen data with high precision.
\end{abstract}

\section{Introduction}

Deep learning approaches, especially neural network-based ones, have been widely employed and proven effective in sentiment analysis tasks \cite{rosenthal2017semeval, nakov2016semeval}. These performance improvements, however, have come at the cost of model transparency and accountability \cite{inkpen2019human}. Many times, deep models are being used as black-box tools by users, even without knowing the model's peculiarities as well as limitations \cite{IBM2018}. In that sense, users can be easily exposed to impropriety or error predictions made in run time, and thus lose trust towards the sentiment classification system.

Traditional evaluation metrics, such as accuracy and F1-score, can explain the predictive performance of a sentiment model. However, their explanations are from an overall and reactive perspective, as they fail to provide insights into the details on when and why the sentiment models fail in run-time \cite{nushi2018towards}. Manual error analysis or heuristics-based error analysis are also common methods for error identification, however, both of them requires either human intervention or domain knowledge, either in the form of labeled data \cite{stymne2011blast} or declarative information (e.g., heuristics or knowledge bases) \cite{bassil2012ocr}. However, labeling instances can be time and effort consuming, and pre-defined knowledge applicable to a specific model is difficult to get. To address this concern, researchers and practitioners have recently raised the need for developing more proactive error detection mechanisms to increase the accountability of the sentiment classification systems while in use. Such accountability mechanisms should be able to identify and measure prediction errors as well as to provide prompt notifications and rectification to the users \cite{crawford2016ai}.

With this challenge in mind, we introduce in this study an explainable error detection framework with human-in-the-loop for sentiment analysis task. Specifically, as shown in Figure 1, given a pre-trained black-box sentiment model, the error-detection framework first analyzes local feature contributions through a data perturbation process. Next, the local feature contributions are aggregated for global-level feature contributions. Later, humans are brought into the loop to assess the relevance of the top ranked global features to the target sentiment classes, and report errors if any. An erroneous score is calculated based on both global and local features. Instances exhibiting erroneous scores above a specific threshold are flagged as problematic predictions. We demonstrate the error detection framework on two sentiment test datasets. From experimental results, we notice high error detection precision of the proposed framework.

\begin{figure}[!htbp]
\centering
  \includegraphics[width=\linewidth]{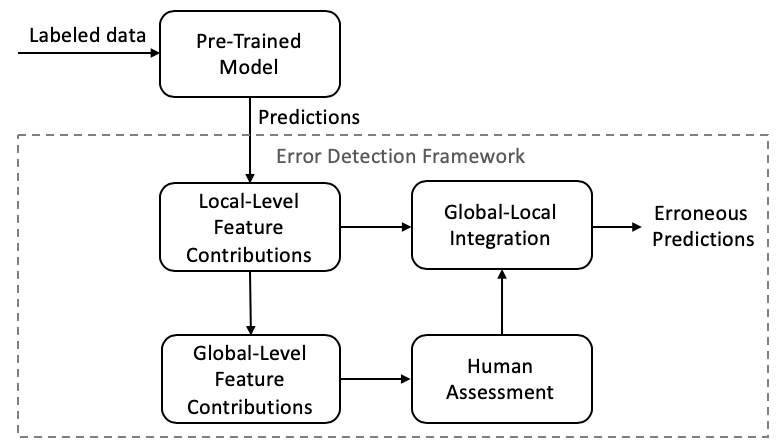}
  \caption{Overview of the explainable error detection framework.}~\label{fig:figure1}
\end{figure}

Our contributions are fourfold: First, we present a high precision error detection framework for sentiment analysis task, which can proactively notify users with prediction errors in run-time. Second, the proposed framework calculates the likelihood of concerns on model correctness in even unseen cases. Third, erroneous predictions are identified based on explainable features which allow the users to easily understand why a prediction fails. Fourth, the proposed error detection framework requires little human effort in error detection by labeling on the global feature level, rather than local instance level.

\section{Related Work}

Error analysis and detection is important to building accountable AI models, as they allow individuals to understand when and how predictions fail \cite{nushi2018towards}. The most intuitive error analysis method is to evaluate the algorithms on a broad set of performance metrics, such as, sensitivity and specifity analysis \cite{harper2009facts}. But their explanations are from an overall and reactive perspective, as they fail to provide insights on each specific instance level on unseen data. Manual error analysis is another common approach, whereas it requires significant human efforts and time, and is not easy to scale. By taking prior knowledge, such as semantic context, into consideration, heuristic-based method brings context-based errors into practice \cite{bassil2012ocr}. Comparing to the manual error analysis, heuristic-based method requires no human intervention. However, pre-defined heuristics applicable to a specific model is difficult to get. Uncertainty sampling \cite{settles2012active} based on a model's confidence scores to identify and label potential prediction errors. Although no human effort is needed in this approach, its performance is not always reliable.  In addition, a series of methods have recently been introduced to identify model errors in a human-machine interactive manner \cite{fiebrink2011human, chen2018anchorviz, nushi2018towards}. By adding humans into the model evaluation step \cite{fails2003interactive}, the proposed methods allowed the users to improve the model performance iteratively, by identifying model errors, providing new training data rewardingly, and retraining the model. Common limitations of these human-in-the-loop based methods are that, they require human labeling on the instance level, which can be labor-intensive, and can not be easily generalized to unseen data.

\section{Methods}

In this section, we present our framework to detect errors in sentiment predictions with human-in-the-loop in detail. Given a blackbox sentiment model and a set of unseen test data, the proposed method runs over the following four steps: 1) ``local-level feature contributions'' module quantifies the feature contributions to the prediction of each individual target instance. 2) ``global-level feature contributions'' module characterizes the general effect of a feature to the overall prediction across all instances.  3) ``human assessment'' module
brings human into the loop of error detection by allowing them to manually label on an interpretable feature-level, instead of instance-level, to save the labeling efforts. 4) ``global-local integration'' module quantifies the erroneous probabilities of instance-level predictions made by the model. With the erroneous probabilities, the framework can send users with failure alerts in prediction run-time on unseen sentiment data.

\subsection{Local-Level Explainable Feature Contributions}

Local-level feature explanations refers to the interpretations used to justify why the model made a specific decision for a single instance. Many existing approaches \cite{lundberg2017unified, lakkaraju2017interpretable} can be adopted for local interpretations. Among the many existing explanation-generating methods, we adopted LIME \cite{ribeiro2016should} as an example way for achieving local-level feature importance in the proposed framework. LIME relies on random perturbation to artificially generate datasets around an instance and then using the generated dataset to train local linear interpretable models for single instance level explanations.

\begin{figure}[!htbp]
\centering
  \includegraphics[width=\linewidth]{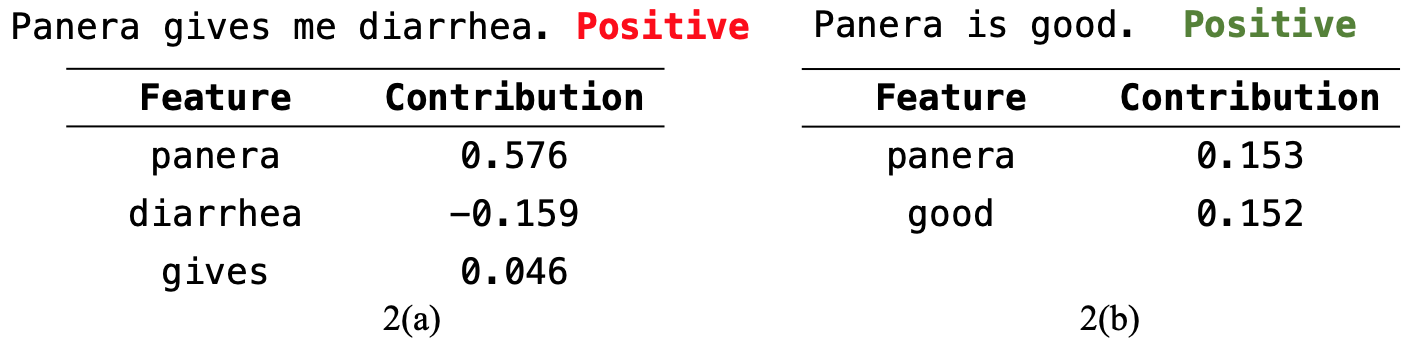}
  \caption{Example of LIME generated local-level
feature contributions. }~\label{fig:figure2}
\end{figure}

In the case of sentiment analysis, we chose unigrams as the explainable feature for LIME, as it is the smallest unit of a text snippet carrying sufficient information that can be reasonably interpreted by human. We implemented linear regression as the base model in LIME and applied it on our dataset, and ranked the explainable features based on their derived coefficients. An example of the local-level feature contributions produced by LIME in our case is presented in Figure ~\ref{fig:figure2}, where the pre-trained model's prediction for sentence ``Panera gives me diarrhea.'' in ~\ref{fig:figure2}(a) is ``positive''. The unigram feature ``panera'' contributes positively to the positive prediction with a magnitude of 0.576, whereas ``diarrhea'' contributes negatively to the positive prediction with a magnitude of 0.159. By observing the local-level feature contributions, we can tell that the model makes a prediction error by considering the word ``panera'' as a significant indicator of the positive polarity, and thus led the model to assign a positive label to the negative sentence.

\subsection{Global-Level Explainable Feature Contributions}

\begin{figure}[!htbp]
\centering
  \includegraphics[width=\linewidth]{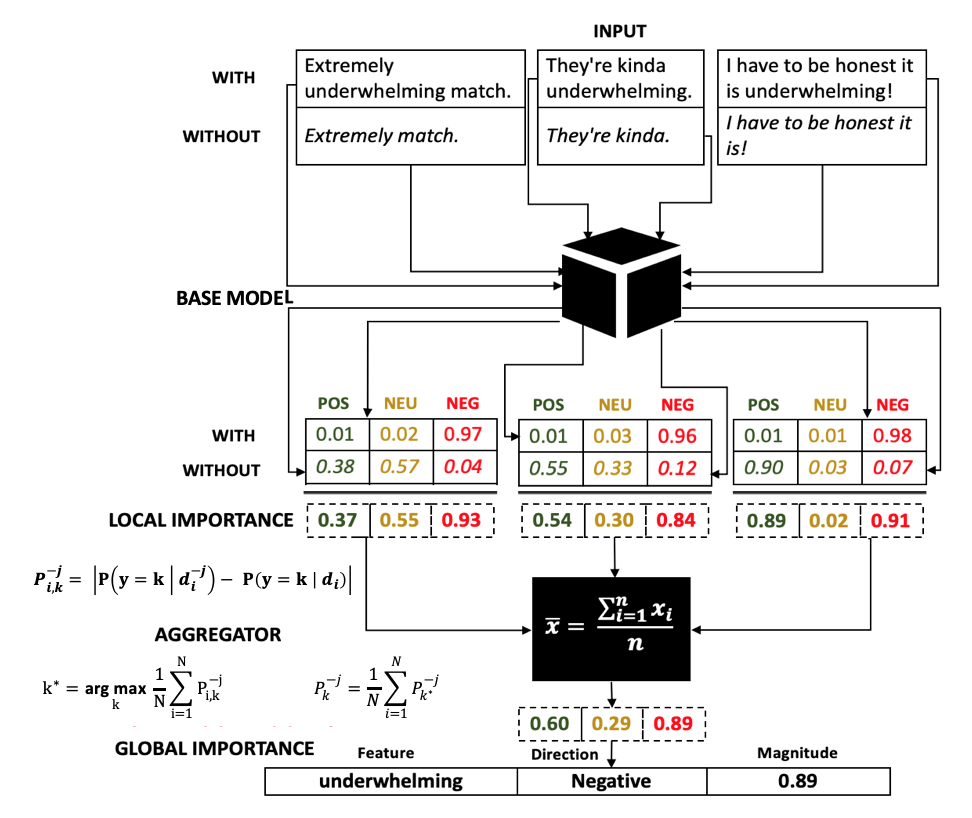}
  
  \caption{Perturbation-based method for achieving global feature contributions. }~\label{fig:figure3}
\end{figure}

Global-level explainable feature contribution demonstrates how each explainable feature affects the model's prediction with regard to the whole training samples, instead of individual instance level predictions \cite{molnar2019}. Achieving global-level feature contributions \cite{molnar2019, letham2015interpretable, arguello2009classification} can help extract more distilled knowledge for less human efforts and can thus facilitates user's understanding of the whole prediction logic behind the model. In the proposed framework, we ran the perturbation-based analysis first on the local-level for all training samples (although it could be on testing as well \cite{molnar2019}) by masking individual feature $j$, one at a time, from each data instance $d_i$, $i \in \{0,1,…,N\}$, which contains feature $j$. We then calculated the absolute changes in the model's prediction probabilities associated with each class label $k \in \{0,1,…,K\}$ as:
\[P_{i, k}^{-j} = |P(y = k | d_i^{-j}) - P(y = k | d_i)| \]
where $P(y = k | d_i)$ is the pre-trained model's prediction probability with feature $j$, and $P(y = k | d_i^{-j})$ is the probability without $j$. We denoted the $P_{i, k}^{-j}$ as feature $j$'s local importance associated with class $k$ to data instance $d_i$. With all $N$ instances containing feature $j$, we finally aggregated the local level importance of $j$ to all $d_i$ to a global level as:

\[k^* = \argmax_k \frac{1}{N}\sum_{i=1}^{N} P_{i, k}^{-j}\]

and denoted class label $k^*$ with the maximum average probability change as the direction of feature $j$'s contribution, and the associated $P_k^{-j}$ as its contribution magnitude, where:
\[P_k^{-j} = \frac{1}{N}\sum_{i=1}^{N}P_{k^*}^{-j} \]
The global importance measurement can be viewed as an aggregation of the local contributions. The underlying assumption behind this method is that a feature is important, if removing it can change the prediction probability significantly. Using the unigram feature ``underwhelming'' as an example, Figure ~\ref{fig:figure3} shows how the corresponding feature magnitude and direction are achieved by using the proposed method. Features were ranked in descending order according to their derived global contribution magnitudes. This would allow us next to show the more important features earlier to the human assessors, so as to help them identify the most significant errors in the shortest time. 

\subsection{Human-in-the-loop Assessment on Global Errors}
The human-in-the-loop assessment module requires human assessors to screen the top N globally contributing features learnt from the previous step with regard to their predicted sentiment labels. Compared with labeling on an instance basis, feature level annotation could be a lot more efficient. Although the top contributing features are not necessarily error-prone, they are more likely to affect the model's overall performance and allow us to zoom into erroneous model predictions in a more efficient manner. 

In the proposed algorithm, we asked the human assessors to label on an unigram basis, given the following considerations: First, labeling on unigrams may lead to more generalized outputs as compared to labeling on bigrams/trigrams. Assuming that the same number of erroneous features were identified by the human assessors on both the unigram and the bigram/trigram levels, more potentially erroneous instances containing the identified unigrams could be found as compared to instances containing the identified bigrams/trigrams, as they are often too sparse or too content specific. Second, labeling on unigrams may be less time and effort intensive, as unigrams tend to be more interpretable to human assessors. If we choose any adjacent words as our bigrams/trigrams, under many circumstances we would not get meaningful phrases, on which labeling could be difficult. Third, existing work on constructing polarity lexicons with manual annotation decisions \cite{mohammad2013crowdsourcing, rouces2018generating} were successfully performed on the unigram basis.

To be more specific, in this step, human assessors were asked to rate the correctness of the globally learnt contribution directions (positive, negative or neutral) of the top N unigrams. Given that the top contributing unigrams were not context-specific or sense-disambiguated, we followed the same heuristic as in \citet{rouces2018generating} by showing only the definition of the first sense in WordNet \cite{miller1995wordnet} to the assessors for annotation purpose, as the first sense is by design the most common meaning of a word.  For unigrams that can not be found in WordNet, we showed the top definition from Urban Dictionary \footnote{https://www.urbandictionary.com} instead. An example
annotation task would be: for unigram feature ``panera'', we first displayed the word itself to the assessor, followed by its first definition in Urban Dictionary, and its contribution direction of being ``positive'' as learnt from the global feature contribution step. We then asked the assessors to rate their agreement on ``panera'' with the definition ``A clean, upscale chain of restaurants primarily located on the eastern coast of America'' as of ``positive'' polarity on a 5-points Likert scale (1: Strongly Disagree, 2: Agree, 3: Neutral, 4: Disagree, 5: Strongly Disagree). As acquiring assessment from experts would be expensive, assessment can be done in a crowd-sourced manner.

\subsection{Global-Local Integration}
The erroneous features recognized on the global level could indeed help identify problematic predictions on the unlabeled instances. However, flagging error occurrence on individual instance level only based on these problematic features may also be unreliable. As shown in Figure 2, noticing ``panera'' being incorrectly learned as ``positive'' could help us accurately identify the wrong prediction of sentence 2(a). However, its erroneous impact on sentence 2(b) is disguised by the existence of the other positive feature ``good'', which were actually learned correctly on the global level.

To more accurately identify the problematic predictions on unlabeled instances, in this step we proposed a measurement metric called the local erroneous score $e$, to determine the relative impact of the global erroneous features on the local level. $e$ was calculated as a normalized version of the accumulated error contributions induced by the globally identified problematic features:

\[e = \frac{\sum_{i=1}^{m}c_{j}^{*}}{\sum_{i=1}^{n}c_{i}^{+}} \]

where $c_j^* \in [-1,1]$ represents the local contribution of the erroneous feature $j$ on the specific instance, and $m$ indicates the total number of erroneous features identified from the global perspective. $c_i^+ \in (0,1]$ represents the local contribution of the feature $i$, whose contribution direction is the same as the final prediction. $n$ specifies the total number of positively contributed features. The local erroneous score e has the value between $-\infty$ to 1. By applying the proposed equation on the two examples as shown in Figure 2, we can see that sentence 2(a) derived a much higher local erroneous score of 0.926, than sentence 2(b) of score 0.502. This demonstrated the effectiveness of the proposed measurement in terms of error detection. A pre-defined threshold $\tau$ is set by the user, and only instances with $e > \tau$ would be retrieved as problematic predictions.

\section{Experiment Settings}

To test our error detection framework, we first create a three-class sentiment classifier, and later treat it as a black-box ``pre-trained'' model for error detection. We implemented the classifier using a replication of the multichannel CNN model introduced by \citet{kim2014convolutional}, although any algorithm can be applied here as the black-box pre-trained model. The training data contains 2,265,413 positive, 2,704,587 negative, and 2,297,426 neutral cases. They were collected from various sources, ranging from high-quality human-labeled instances to pseudo-labeled instances annotated using emoji or hashtag based indicators \cite{novak2015sentiment}. We evaluated the model on the test dataset of SemEval2016 Task 4 Subtask A \cite{nakov2016semeval}. Our model achieved a $F_1^{PN}$ of 0.345, and a three-class prediction accuracy of 0.463, which is comparable to many of the SemEval2016 participation systems. 

We calculated the local and global feature contributions using the training dataset and performed the human-in-the-loop assessment on Figure Eight \footnote{https://www.figure-eight.com/}. We extracted the top 2,000 non-neutral features with the highest global contribution magnitudes to human assessors to determine their global correctness. We chose only non-neutral features for error assessment as polarity errors, such as predicting positive as negative or vice versa, can greatly impact user's trust towards the model, and we want to focus more on such extreme cases for error detection. 5 unigrams were shown in 1 annotation page and gold questions (easy questions with known answers, e.g. ``happy'' with the definition ``enjoying or showing or marked by joy or pleasure'' as ``positive'' polarity) were embedded on each page for quality check. For each unigram feature, we recruited in total 5 assessors who had to be native English speakers, with the highest level of experience. For all 2,000 unigram features, we collected in total 10,155 judgements within 1 hour. Among them only 155 (1.5\%) were from untrusted assessors, who have been excluded during the annotation process. This indicated the relative easiness of feature labeling for sentiment task for even non-expert assessors. We obtained the annotations for 1,725 out of the 2,000 assessed unigrams and converted them into binary cases (agree or disagree, 86.25\% inter-rater agreement), where at least 3 assessors agree on the same answer. Among them 161 were found to be wrongly learned by the pre-trained black-box model. Global-level features include ``kashmir'', ``midterm'', ``dems'', ``netflix'' was being wrongly predicted by the model as ``negative'', whereas ``wingstop'', ``panera'', ``minister'', ``popeyes'' as ``positive''. All 161 global-level problematic features would then be passed to the next step to guide the instance-level error identification.

To assess the method's effectiveness, we applied the proposed error detection framework first on the test dataset of SemEval2016 Task 4 Subtask A. We found 932 instances containing at least one of the globally identified problematic features. Thus, we only calculated the local erroneous score $e$ for the 932 cases.  Given that the SemEval dataset only covers a subset (60/161) of the erroneous unigrams, we prepared another dataset customized just for better understanding of the precision of the proposed framework. Specifically, we adopted the 161 problematic unigrams as search keywords to collect tweets using Twitter Search API. For each keyword we collected up to 50 most recent tweets. We cleaned the collected dataset by removing duplicate tweets and tweets with URL, assuming that they have a higher probability of being spam. In total, we collected 3,111 instances in this customized Twitter testing dataset. 

For both datasets, we extracted all instances with $e > \tau$. For the self-collected Twitter data, we acquired the ground truth labels from the crowd on Figure Eight. We reported the precision of the proposed method at different settings of the threshold $\tau$, to understand its impacts on the framework's performance.

We adopted uncertainty sampling as the baseline to evaluate the performance of the proposed error detection framework. We adopted the least confidence as the measurement in this experiment, which is based on the difference between the most confident prediction and 100\% confidence. In other words, for a three-class classification task, the model is most unconfident when having the maximum prediction probability around 0.33. We applied uncertainty sampling just on the complete SemEval test data, assuming no filtering at all was applied on the original dataset. We extracted instances with the lowest prediction confidence as under-trained cases and compared the precision@K for both the baseline method and the proposed method.

\section{Results}
In Figure 4, we plotted the error detection precision for the proposed method on both datasets, along with varied thresholds of $\tau$ ranging from 0 to 0.4. We set the upper bound $\tau$ to 0.4 instead of 1, since only very limited number of instances (sizes with no or little statistical meaning) were detected for the SemEval dataset when $\tau \geq 0.5$.

\begin{figure}[h]
\centering
  \includegraphics[width=\linewidth]{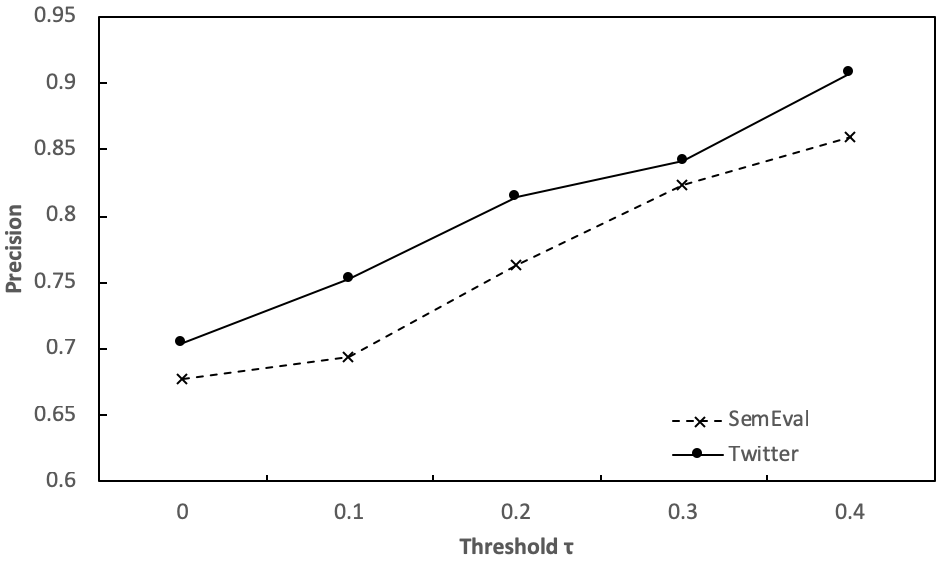}
  \caption{Precision for the error detection framework with varied threshold settings. }~\label{fig:figure4}
\end{figure}

As can be seen from Figure 5, even when the threshold $\tau$ was set to a low score of 0, the proposed error detection framework can still achieve relatively high precision scores for both datasets. Specifically, for the SemEval dataset, with $\tau = 0$, the proposed framework indicated 48.9\% of all prediction instances as erroneous predictions. Among these flagged instances, 67.7\% were proved to be truly problematic according to the ground truth labels. The same pattern was also noticed for the self-collected Twitter dataset, when $\tau = 0$, the proposed method detected in total 57.9\% suspicious predictions, and 70.5\% of them were proved to be truly problematic. Looking further, we observed that the error detection framework became even more precise, as we incrementally increased the value of $\tau$. It reached the highest precision of 85.9\% for the SemEval and 90.8\% for the Twitter dataset when $\tau = 0.4$. But obviously these increases in precision were achieved with a trade-off of the degradation in the number of detected problematic predictions.

Considering our ultimate goal of altering users of potential prediction errors, we next compared the proposed error detection framework with uncertainty sampling based on precision@K. Precision@K
is a widely adopted evaluation metric in information retrieval tasks. It is being defined here as the proportion of identified erroneous cases that are real errors in the top $K$ retrieved results. We chose K ranging from 100 to 400, as only 455 predictions are being identified as problematic with $\tau \geq 0$. Table 1 shows the precision@K for both uncertainty sampling and the proposed approach. 

As can be noticed in Table 1, when alerting the users with the top 100 identified error predictions, the proposed error detection method with human-in-the-loop showed significant performance advantage over uncertainty sampling with a 0.110 precision gap. Such advantages gradually decreased as $K$ became larger (more relaxed $\tau$). When $K = 400$, the prediction probability threshold for uncertainty sampling equaled to 0.393, and the precision gap between the two methods decreased to 0.006. 

\begin{table}
\centering
\begin{tabular}{lcc}
\hline
\textbf{\textit{K}} & \textbf{\textit{Uncertainty}} & \textbf{\textit{Human-in-the-loop}}\\
\hline
100 & 0.710 & 0.820 \\
200 & 0.685 & 0.805 \\
300 & 0.686 & 0.750 \\ 
400 & 0.692 & 0.698 \\\hline
\end{tabular}
\caption{Precision@K for uncertainty sampling and the proposed error detection method with human-in-the-loop.}
\label{tab:accents}
\end{table}


In addition to precision, we were also interested in knowing if the proposed method was able to catch errors that can not be detected by the uncertainty sampling baseline. To achieve this goal, we plotted in Figure 5 the distribution of the prediction probabilities of the erroneous cases identified by the proposed framework when $\tau = 0.4$. From Figure 5, we found that about 40\% of the erroneous instances detected by the proposed method were associated with prediction probabilities of larger than 0.7. In other words, this means that the model is quite confident about those predictions and uncertainty sampling would hardly treat them as potential errors. In that sense, we conclude that the proposed framework is able to detect errors even when the prediction confidence is high. 

\begin{figure}[h]
\centering
  \includegraphics[width=\linewidth]{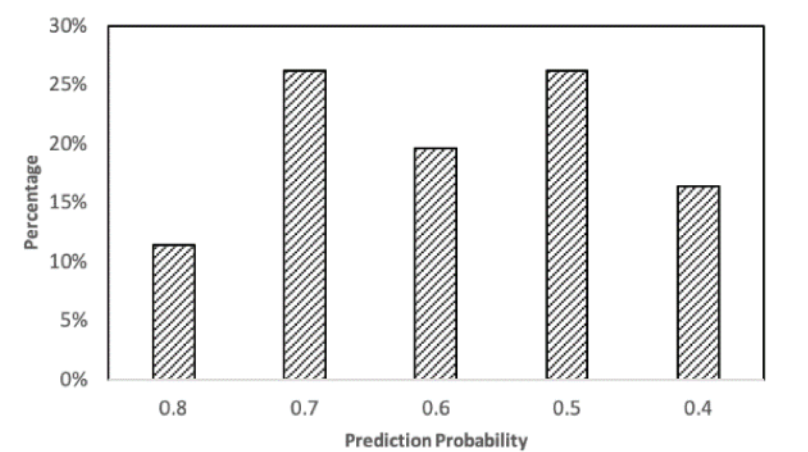}
  \caption{Uncertainty probability distribution for erroneous cases when $\tau = 0.4$. }~\label{fig:figure5}
\end{figure}

\section{Discussion and Conclusion}
Our work was motivated by the practical concerns of not knowing the limitations or potential errors of a sentiment model prior to deployment, and not being able to notify or even rectify when erroneous predictions were made once deployed. Driven by this demand, in this paper, we presented a framework for identifying prediction errors in an interpretable manner for sentiment analysis tasks. We validated the proposed error detection framework on two different datasets. Results showed that the proposed method can identify problematic model predictions with high precision, which is critical to continuous model refinement.  While comparing the proposed approach with the baseline, we noticed that our method can also be adopted as a selective sampling approach, in addition to uncertainty sampling. Besides, given that the proposed error detection framework can be applied on unseen data without ground truth, it can proactively notify users about possible erroneous decisions made by the model in prediction run-time. 

In addition to its effectiveness, the proposed error detection framework can also be easily explained to the users. Globally, the proposed framework allows the users to understand the overall contribution of a word to the final predictions made by the black-box sentiment model. Besides, as demonstrated in our results, global-level contributions can also be useful for identifying potential bias existed in the model. For instance, we noticed that ``netflix'', ``dems'', and ``palestine'' were being learned as ``negative'' in our pre-trained sentiment model. While integrating global-level problematic features with local-level predictions, the proposed erroneous score enables users to know why that specific prediction could be wrong or biased and how much the problematic global feature contributed to the erroneous or biased prediction. Certainly, more work is needed on how the proposed framework can be generalized to the task of bias detection.

Furthermore, the proposed framework efficiently integrated human into the loop of model validation and refinement. Comparing with the previous methods, our approach allows the human annotators to label on the explainable feature level, rather than on the instance level, which can significantly save their time and effort. Regarding the annotation quality of the feature level labeling, our results showed that even non-expert crowd workers can accurately finish the assessment tasks with high inter-rater agreement in a very short period of time. 

Finally, our work comes with certain limitations. One of them is the relatively small number of global features (2,000) that were labeled by the human assessors in this work. To some extent, this limited us from evaluating the presented error detection framework from more angels other than precision, although precision is the most important measurement for error detection. Annotations on larger scales will be conducted in later studies and the effectiveness of the proposed framework will be evaluated from more angels. Besides, future works will also be conducted on investigating how these identified erroneous instances or features could be used for further fixing or debugging the pre-trained models.



\balance
\bibliography{anthology,custom}
\bibliographystyle{acl_natbib}

\end{document}